\documentclass[sigconf]{acmart}
\AtBeginDocument{%
  }

\setcopyright{acmlicensed}
\copyrightyear{2025}
\acmYear{2025}
\acmConference[ICMR '25] {Proceedings of the 48th International ACM ICMR Conference on Research and Development in Information Retrieval}{ June 30--July 3, 2025}{Chicago, IL, USA.}
\acmBooktitle{Proceedings of the 48th International ACM ICMR Conference on Research and Development in Information Retrieval (ICMR '25), June 30--July 3, 2025, Chicago, IL, USA}
\acmISBN{979-8-4007-1592-1/25/07}
\acmDOI{10.1145/XXXXXX.XXXXXX}

\usepackage{multicol}
\usepackage{multirow}
\usepackage{graphicx}
\begin{document}

\title{Audio-Driven Talking Face Video Generation with Joint Uncertainty Learning}

\author{Yifan Xie}
\authornote{Both authors contributed equally to this research.}
\email{ivanxie416@gmail.com}
\orcid{1234-5678-9012}
\affiliation{%
  \institution{Guangdong Laboratory of Artificial Intelligence and Digital Economy (SZ)}
  \city{Shenzhen}
  % \state{Guangdong}
  \country{China}
}

\author{Fei Ma}
\email{mafei@gml.ac.cn}
\authornotemark[1]
\affiliation{%
  \institution{Guangdong Laboratory of Artificial Intelligence and Digital Economy (SZ)}
  \city{Shenzhen}
  \country{China}}

\author{Yi Bin}
\email{yi.bin@hotmail.com}
\affiliation{%
  \institution{Tongji University}
  \city{Shanghai}
  \country{China}}

\author{Ying He}
\email{heying@szu.edu.cn}
\affiliation{%
  \institution{Shenzhen University}
  \city{Shenzhen}
  \country{China}}

\author{Fei Yu}
% \authornotemark[2]
\authornote{Corresponding author.}
\email{yufei@gml.ac.cn}
\affiliation{%
  \institution{Guangdong Laboratory of Artificial Intelligence and Digital Economy (SZ)}
  \city{Shenzhen}
  \country{China}}

% \affiliation{%

% \affiliation{%

% \affiliation{%

\renewcommand{\shortauthors}{Xie et al.}

\begin{abstract}
Talking face video generation with arbitrary speech audio is a significant challenge within the realm of digital human technology. 
The previous studies have emphasized the significance of audio-lip synchronization and visual quality. 
Currently, limited attention has been given to the learning of visual uncertainty, which creates several issues in existing systems, including inconsistent visual quality and unreliable performance across different input conditions.
To address the problem, we propose a Joint Uncertainty Learning Network (\textit{JULNet}) for high-quality talking face video generation, which incorporates a representation of uncertainty that is directly related to visual error.
Specifically, we first design an uncertainty module to individually predict the error map and uncertainty map after obtaining the generated image. The error map represents the difference between the generated image and the ground truth image, while the uncertainty map is used to predict the probability of incorrect estimates.
Furthermore,
to match the uncertainty distribution with the error distribution through a KL divergence term,
we introduce a histogram technique to approximate the distributions.
By jointly optimizing error and uncertainty, the performance and robustness of our model can be enhanced.
Extensive experiments demonstrate that our method achieves superior high-fidelity and audio-lip synchronization in talking face video generation compared to previous methods.
\end{abstract}





\maketitle

\section{Introduction}
With the advancement of the audio-visual industry, the audio-driven talking face video generation has become increasingly important in the computer vision and multimedia technology field. This research topic has gained prominence in recent years and has the potential to be integrated into various applications such as virtual avatars~\cite{cheymol2023beyond}, online meetings~\cite{kagan2023zooming}, and film dubbing~\cite{ma2024generative,luo2024codeswap,ma2024review}.
However, there has been limited attention given to the learning of visual uncertainty in talking face generation. To address the problem, we propose a novel approach that incorporates uncertainty estimation into the generation process, inspired by the combined consideration of epistemic and aleatoric uncertainty in computer vision tasks, where uncertainty learning~\cite{kendall2017uncertainties} has proven beneficial for improving model performance and reliability.

\begin{figure}[t]
\begin{center}
\includegraphics[width=1.0\linewidth]{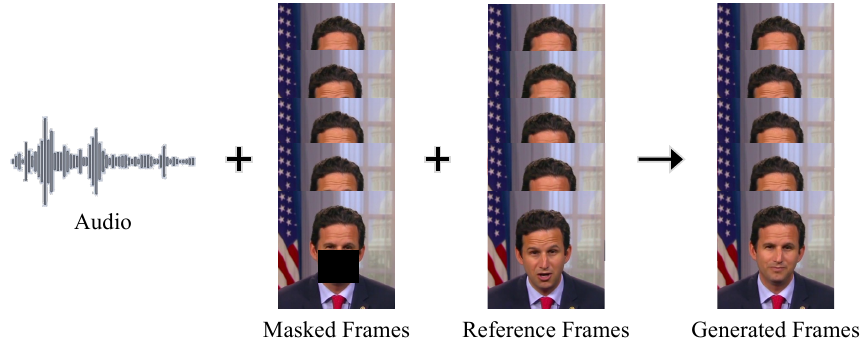}
\end{center}
% \vskip -10pt
\caption{The process of audio-driven talking face video generation. Given the audio clip, multiple masked frames, and reference frames, our method can generate a high-fidelity talking face video.}
\label{figure1}
% \vskip -10pt
\end{figure}

Given a sequence of source videos and a driven audio clip, the objective is to produce a talking face video that matches the source identity, perfectly synchronized with the driven audio, while providing reliable uncertainty estimates for the generated content. Previous works have demonstrated that uncertainty is highly correlated with prediction errors, motivating us to train an uncertainty network that models the distribution of true errors in generated image pixels.
Existing methods for generating talking face videos can be divided into two categories: person-specific and person-generic, depending on the training paradigm and data requirements. Person-specific methods~\cite{guo2021ad,tang2022real,li2023efficient,peng2023synctalk,xie2025pointtalk} are capable of generating realistic talking face videos, but they require re-training or fine-tuning with videos of the target speaker. This can be problematic in real-world scenarios where the target speaker's videos may not be accessible. Therefore, the more significant and challenging task in this field is to learn how to generate person-generic talking face videos~\cite{prajwal2020lip,zhong2023identity,wang2023seeing,zhang2023dinet,shen2023difftalk}.

For person-generic talking methods, Wav2Lip~\cite{prajwal2020lip} stands as the pioneering method in incorporating a lip synchronization expert to ensure accurate lip movements, while TalkLip~\cite{wang2023seeing} utilizes a lip-reading expert to enhance the clarity and precision of these movements. DINet~\cite{zhang2023dinet} focuses on preserving high-frequency textural details by spatially deforming reference images. IP-LAP~\cite{zhong2023identity} introduces a novel two-stage framework for generating identity-preserving talking faces by incorporating landmark and appearance priors based on audio input. Building on these methods, our JULNet aims to tackle person-generic talking face video generation with uncertainty awareness. Specifically, we utilize audio clips and multiple reference images as guidance to complete the lower-half mouth region of the speaker's original video, while simultaneously estimating the uncertainty of our predictions to enhance the model's robustness and reliability.
The process is depicted in Figure~\ref{figure1}.

% Inspired by the combined consideration of epistemic and aleatoric uncertainty in~\cite{kendall2017uncertainties}, this study introduces innovative loss functions that allow for uncertainty estimation in pixel-wise vision tasks. The results obtained from segmentation and depth regression demonstrate how the main task can benefit from concurrent uncertainty learning. Moreover, it suggests that modeling aleatoric uncertainty is the most effective way, while epistemic uncertainty can be reduced when a large amount of data is available. Consequently, our focus is primarily on aleatoric uncertainty. Previous works~\cite{duggal2019deeppruner,chen2023learning} have proved that uncertainty is highly correlated with error. Our motivation is to train an uncertainty network that produces outputs following the same distribution as the true errors of the generated image pixels. By jointly optimizing error and uncertainty, we can enhance the performance and robustness of the model.

In existing talking face generation methods, a key challenge is the lack of reliability measures for the generated content. Inspired by uncertainty estimation in computer vision tasks~\cite{kendall2017uncertainties}, we introduce uncertainty maps to explicitly model the prediction confidence of our network. These uncertainty maps help identify regions where the generation might be unreliable, allowing the model to better handle challenging cases such as extreme head poses or complex lip movements. Following previous findings that uncertainty is highly correlated with prediction errors~\cite{chen2023learning}, we train our network to estimate uncertainty that reflects the true error distribution in generated pixels, leading to improved image quality and more accurate audio-visual synchronization in challenging scenarios.

In this paper, we propose a novel joint uncertainty learning network, named JULNet, to achieve high-quality talking face video generation.
To the best of our knowledge, the proposed JULNet is the first method to introduce uncertainty learning for this task.
Specifically, given the audio clip, masked image and multiple reference images, features are first extracted by respective encoders, and then these features are concatenated together to generate an image via a generator. Additionally, a discriminator is utilized to enhance the quality of the generated images.
After obtaining the generated image, we introduce an uncertainty module to predict the error map and uncertainty map separately.
The error map represents the difference between the generated image and the ground truth image, while the uncertainty map is used to predict the probability of incorrect estimates. To match the uncertainty distribution with the error distribution, we develop a histogram technique to approximate the distributions via a Kullback-Leibler (KL) divergence term.
Furthermore, we integrate the perception loss and the lip-sync loss to further enhance the performance of our method.
Extensive experiments demonstrate that our JULNet generates realistic talking face videos, characterized by high visual quality and precise audio-lip synchronization.

In conclusion, our contribution can be summarized as:
\begin{itemize}
  \item We propose a joint uncertainty learning network called JULNet for talking face video generation, which improves the performance by incorporating a representation of uncertainty that is directly related to visual error.
  \item We introduce an uncertainty module to predict the error map and the uncertainty map. The error map represents the differences, while the uncertainty map indicates the probability of incorrect estimates.
  \item We integrate a histogram technique to approximate the error distribution and the uncertainty distribution, and utilize a KL divergence term to match the distributions.
\end{itemize}

\section{Related Work}

\subsection{Talking Face Video Generation}
% The study of generating talking face videos began in the 1990s with the goal of mapping acoustic features to realistic facial movements that are synchronized with audio.
% Traditional methods~\cite{yehia1998quantitative,brand1999voice} utilize hidden Markov models to predict facial motions. For example, Brand~\cite{brand1999voice} predicts facial motions by learning hidden Markov models from time-aligned audio-visual streams with minimized entropy. 
% In recent years, deep learning has also made significant advancements in this area. These advancements have benefited the task of facial motion prediction. Specifically, 
% the majority of these works can be classified as either person-specific~\cite{suwajanakorn2017synthesizing,chen2020talking,ji2021audio,guo2021ad,tang2022real,li2023efficient,peng2023synctalk} or person-generic~\cite{prajwal2020lip,zhong2023identity,wang2023seeing,zhang2023dinet,shen2023difftalk,ma2023dreamtalk} methods.
\subsubsection{Person-Specific Talking Face Video Generation}
% \noindent \textbf{Person-Specific Talking Face Video Generation.}
Conventional person-specific methods~\cite{chen2020talking,ji2021audio} often utilize 3D Morphable models (3DMM)~\cite{blanz2023morphable} to generate talking face videos for specific individuals. 
These methods rely on the 3D structural modeling of the head to create more realistic-looking videos. However, the use of intermediate representations can cause errors to accumulate.
The recent emergence of neural radiance fields (NeRF)~\cite{mildenhall2021nerf} has revolutionized the field of 3D head structure problems in talking face video generation. 
AD-NeRF~\cite{guo2021ad} stands as the pioneering method in utilizing NeRF for this task, but it faces challenges with slow training and inference speeds. 
To address these issues, RAD-NeRF~\cite{tang2022real} incorporates Instant-NGP~\cite{muller2022instant}, resulting in significant improvements in both visual quality and efficiency. 
ER-NeRF~\cite{li2023efficient} introduces a triple-plane hash encoder aimed at eliminating empty spatial regions and generating region-aware conditional features using an attention mechanism.
SyncTalk~\cite{peng2023synctalk} enhances the fidelity of audio-driven talking face videos through precise synchronization of facial identity, lip movements, expressions, and head poses.
However, in real-world situations, accessing videos of the target speaker for re-training or fine-tuning may not be possible. Therefore, it is more important to develop person-generic methods that can generate talking face videos for speakers who have not been seen before.

\subsubsection{Person-Generic Talking Face Video Generation}
% \noindent \textbf{Person-Generic Talking Face Video Generation.}
There exists a series of methods focusing on person-generic talking face video generation.
Specifically,
Wav2Lip~\cite{prajwal2020lip} incorporates a lip synchronization expert to oversee the accuracy of lip movements, while TalkLip~\cite{wang2023seeing} utilizes a lip-reading expert to improve the clarity and precision of these movements.
DINet~\cite{zhang2023dinet} preserves high-frequency textural details by deforming reference images.
IP-LAP~\cite{zhong2023identity} introduces a two-stage framework that generates identity-preserving talking faces by incorporating landmark and appearance priors based on audio input. 
DreamTalk~\cite{ma2023dreamtalk} lies in the integration of diffusion probabilistic models for generating expressive talking heads, enhancing accuracy and expressiveness without the need for explicit expression references.
% EAT~\cite{gan2023efficient} introduces the emotional adaptation for audio-driven talking head, which efficiently transforms emotion-agnostic talking head models into emotion-controllable ones through parameter-efficient adaptations.
Recently, diffusion models~\cite{ho2020denoising} have been employed to enhance lip-sync and image quality~\cite{shen2023difftalk,tian2024emo}, but they tend to be slow during inference.
Building on these methods, our JULNet aims to address person-generic talking face video generation through joint uncertainty learning.

\begin{figure*}[t]
\centering
\includegraphics[width=1.0\textwidth]{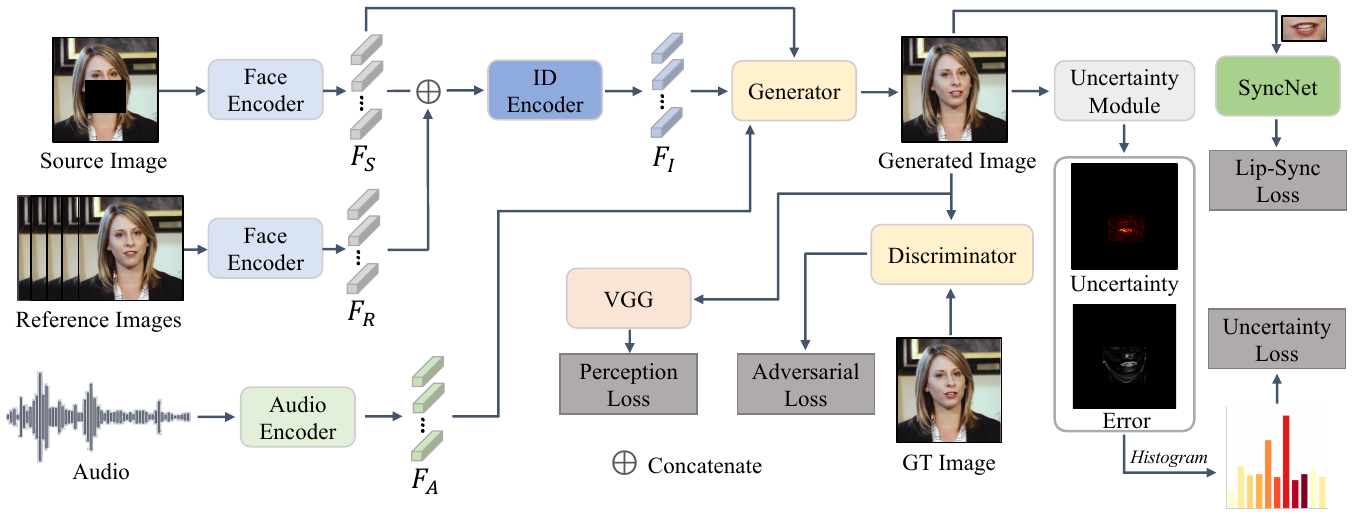} 
\caption{Overall architecture of the proposed JULNet.     
By utilizing audio signals, a masked source image, and a few reference images, the audio encoder, face encoder, and ID encoder are employed to extract the audio feature $F_A$ and the ID features $F_I$. Subsequently, $F_A$ and $F_I$ are concatenated, and the generator is used to synthesize the image. Finally, various optimization processes are carried out on the generated images: (1) a discriminator to differentiate between real and synthesized face images, (2) a pretrained VGG net to minimize the perceptual differences between the generated image and the GT image, (3) a pretrained SyncNet to enhance audio-lip synchronization, and (4) an uncertainty module to generate the uncertainty map and the error map. Additionally, a histogram technique is utilized to approximate distributions and optimize the uncertainty.}
\label{figure2}
\end{figure*}

\subsection{Uncertainty Learning}
Early Bayesian neural networks~\cite{mackay1992practical} are able to quantify uncertainty in shallow networks, but their scalability is limited~\cite{dusenberry2020efficient}. A more practical method for approximating the posterior weight distribution of Bayesian neural networks is the use of deep ensembles~\cite{lakshminarayanan2017simple}, which is widely recognized as the leading technique for uncertainty estimation~\cite{laurent2022packed}. However, deep ensembles come with high computational demands, which has led to the development of alternatives such as deep sparse networks~\cite{liu2021deep} that utilize partially shared weights. Another lightweight alternative is the use of multi-input multi-output networks~\cite{havasi2020training}, which train independent sub networks within a larger network to achieve diversified outputs. However, this approach does require multiple inferences to be made.
% The concept of uncertainty can be easily introduced into reinforcement learning~\cite{zangirolami2024dealing,chiou2024knowledge},
% In the field of computer vision, 
Additionally, there is some related works in the field of computer vision~\cite{poggi2020uncertainty,chen2023learning,zou2022tbrats,cao2024pasco}.
SUB-Depth~\cite{poggi2020uncertainty} thoroughly evaluate uncertainty estimation for self-supervised monocular depth estimation.
SEDNet~\cite{chen2023learning} enhances stereo matching performance through uncertainty estimation.
TBraTS~\cite{zou2022tbrats} generates robust brain tumor segmentation results and dependable uncertainty estimations without excessive computational load.
PaSCo~\cite{cao2024pasco} predicts both voxel-wise and instance-wise uncertainty along panoptic scene completion.
In our work, 
we apply the concept of uncertainty that is directly related to visual error, aiming to generate high-fidelity talking face videos.

\section{Method}
We propose JULNet to generate the realistic talking face videos. The overall architecture of JULNet is illustrated in Figure~\ref{figure2}.
Specifically, JULNet utilizes images as the identity reference and audio signals as the lip movement reference. 
% The process is described in Section~\ref{pipeline}. 
After generating videos using the generator, the uncertainty module is used to predict the error map and uncertainty map. The error map represents the differences, while the uncertainty map indicates the probability of incorrect estimates. 
% as discussed in Section~\ref{up}. 
Additionally, a histogram technique is employed to approximate the distributions, and a KL divergence term is utilized to match the distributions. 
% Finally, the associated loss functions are integrated in Section~\ref{loss}.

\subsection{JULNet Pipeline}

\subsubsection{Encoders}
% \noindent \textbf{Encoders.}
Extracting highly generalized audio features from the input signals is essential for generating talking face videos. 
The audio encoder encodes phoneme-level embedding and provides it to the generator as a reference for lip movement. 
Firstly, we employ a sliding window strategy and utilize an automatic speech recognition (ASR) module~\cite{amodei2016deep,feng2025unisync,feng2025stftcodec} to predict classification labels for each 20ms audio clip.
These labels are further smoothed using 1D convolutions to extract the final audio features $F_A$.
% The use of audio features instead of regressed expression coefficients~\cite{thies2020neural} or facial landmarks is to prevent potential semantic mismatching issues between audio and visual signals.

The source image is identical to the ground truth image, with the exception that it masks the lower-half mouth region to prevent learning lip movements. Five reference images are randomly selected from the same video sequence to preserve identity information. Afterwards, two different face encoders are used to extract source features $F_S$ and reference features $F_R$ separately. Finally, the features are concatenated and passed through an ID encoder to obtain the final face features, denoted as $F_I$. The ID encoder not only preserves identity information, but also aligns the head pose.

\subsubsection{Video Generation}
\begin{figure}[t]
\begin{center}
\includegraphics[width=1.0\linewidth]{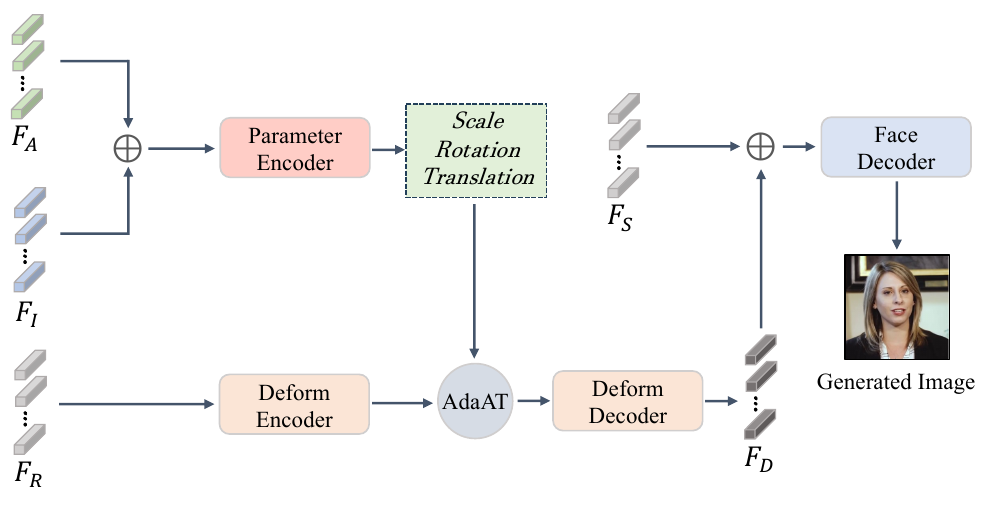}
\end{center}
% \vskip -10pt
\caption{The detailed structure of the generator.}
\label{figure3}
% \vskip -10pt
\end{figure}

% \noindent \textbf{Video Generation.}
The generator synthesis a talking face given the audio features and the face features.
The detailed structure of this module is depicted in Figure~\ref{figure3}.
Following~\cite{zhang2023dinet}, we utilize the AdaAT operator, which can deform feature maps with misaligned spatial layouts by performing feature channel-specific deformations.
It computes various affine coefficients in various feature channels.
Specifically, after concatenating $F_A$ and $F_I$, a parameter encoder is utilized to compute the scale parameter $S$, rotation parameter $R$, and translation parameter $T$. Meanwhile, $F_R$ is used as the input to AdaAT through the deform encoder.
Then, AdaAT employs these parameters to perform affine transformations:
\begin{equation}
\begin{bmatrix}\hat{x}\\\hat{y}\end{bmatrix}=\begin{bmatrix}S*cos(R)&S*(-sin(R))&T_x\\S*sin(R)&S*cos(R)&T_y\end{bmatrix}\begin{bmatrix}x\\y\\1\end{bmatrix},
\end{equation}
where $x/\hat{x}$ and $y/\hat{y}$ denote the pixel coordinates before/after affine transformations. 

After the AdaAT operation, the following deform decoder combines all affine transformations into a single complex spatial deformation, deforming $F_R$ into $F_D$. 
Finally, $F_D$ and $F_S$ are concatenated to generate a face image using a face decoder.

\subsection{Uncertainty Prediction}
\label{up}
Kendall et al.~\cite{kendall2017uncertainties} introduce a network that combines aleatoric uncertainty with epistemic uncertainty. 
It utilizes the negative log-likelihood of the prediction model as the loss function for pixel-wise tasks. 
Based on this, the network can be further optimized to generate an uncertainty distribution that matches the error distribution. Since multi-task learning typically improves the performance of all subtasks, the goal is to minimize the difference between the uncertainty distribution and the actual error distribution and optimize the uncertainty loss.
% 使用L1 loss生成error distribution map，再引出uncertainty map计算公式。

We propose an uncertainty module to predict the error map and the uncertainty map. 
For error map $\epsilon$, each pixel $i$ is calculated by comparing the predicted pixel $p_i$ with the corresponding ground truth pixel $p_i$, using the L1 loss:
\begin{equation}
\epsilon_i = |\hat{p}_i-p_i|.
\end{equation}

Ilg et al.~\cite{ilg2018uncertainty} have shown that in order to predict the uncertainty map and reduce data noise, the predicted uncertainty distribution can be adjusted either as Laplacian or Gaussian, depending on whether the L1 or L2 loss is utilized for error prediction. Therefore, we choose to utilize the Laplacian distribution. The distribution can be described as:
\begin{equation}
p(\epsilon|f_\theta(p,\hat{p}))=\mathrm{Laplacian}(f_\theta(p,\hat{p}),\sigma),
\end{equation}
where $f_\theta$ is an MLP-based network and $\sigma$ denotes the uncertainty parameter. Then, we can construct the pixel-wise uncertainty loss function:
\begin{equation}
\mathcal{L}_{un_1}=\frac1n\sum_{i=1}^n\frac{\epsilon_i}{\exp(\tau_i)}+\frac1n\sum_{i=1}^n\tau_i,
\label{un1}
\end{equation}
where $\tau_i$ is the log of the uncertainty parameter $\sigma_i$ for numerical stability, and $n$ is the total number of the pixels.
The function can be seen as a robust optimization, with the first term being adjusted based on predicted uncertainty, and the second term serving as a regularizer.

\subsection{Distribution Matching}
The process described above enables us to calculate and optimize the uncertainty related to the error. Moreover, it is important to ensure that the error distribution closely aligns with the uncertainty distribution. This alignment allows the error to be accurately represented by the uncertainty, ultimately reducing both the uncertainty and the error.

We utilize the KL divergence to evaluate the dissimilarity between the distributions of error $\epsilon$ and uncertainty $\sigma$. Since the KL divergence is asymmetric, we designate the distribution of error $\epsilon$ as the reference. Therefore, our goal is to minimize the following function:
\begin{equation}
\mathcal{L}_{un_2}=\mathcal{D}(P_\sigma(p))\|Q_\epsilon(p))=\int_0^{p_{max}}P_\sigma(p)\log\frac{P_\sigma(p)}{Q_\epsilon(p)}dp,
\label{kl}
\end{equation}
where $\mathcal{D}$ denotes the divergence between the distributions and $p$ represents the pixel range.
However, minimizing the function requires closed-form expressions for the two distributions. We can assume them to follow Laplacian distribution, but this allows for the uncertainty module to be non-differentiable. Therefore, we integrate a histogram technique to approximate the distributions. The ordinary histogram is also non-differentiable since it is a segmented function over intervals. To address this issue, we design a differentiable statistical distribution for the histogram.

To ensure that the distribution of uncertainty $\sigma$ matches the statistics of the error $\epsilon$, we define a set of bins $B$ for the histogram.
Specifically, we calculate the mean and standard deviation of the error $\epsilon$, referred to as $u_{\epsilon}$ and $s_{\epsilon}$.
Then, we assign $B_0 = u_{\epsilon}$ and $B_m = u_{\epsilon} + \alpha_m s_{\epsilon}$, where $\alpha_m$ represents a temperature coefficient.
Afterward, we establish $m-1$ center points that are evenly distributed on a linear or logarithmic scale between $B_0$ and $B_m$.

To compute histograms for error and uncertainty separately, we start with the bin center points. For each pixel error $\epsilon_i$, we calculate weights $w_j(\epsilon_i)$ for $j$-th bin center point based on their inverse distance:
\begin{equation}
w_j(\epsilon_i)=\lambda_1\cdot\exp(-\frac{(u_\epsilon+\alpha_js_\epsilon-\epsilon_i)^2}{\lambda_2}),
\end{equation}
where $\lambda_1$ and $\lambda_2$ are hyper-parameters. 
For $j$-th bin center point, the sum of the weights of each pixel is counted to get the final histogram representation $H_\epsilon(j)$:
\begin{equation}
H_\epsilon(j)=\frac1n\sum_{i=1}^n\frac{\exp(w_j(\epsilon_i))}{\sum_{j=0}^m\exp(w_j(\epsilon_i))}.
\end{equation}
Similarly, we can obtain $H_\sigma(j)$ in the same way. 
The equation~\ref{kl} can be converted using the KL divergence between the two differentiable histograms:
\begin{equation}
\mathcal{L}_{un_2}=\sum_{j=0}^mH_\epsilon(j)\log\frac{H_\epsilon(j)}{H_\sigma(j)}.
\label{un2}
\end{equation}

\subsection{Loss Functions}

\subsubsection{Uncertainty Loss}
% \noindent \textbf{Uncertainty Loss.}
The uncertainty loss $\mathcal{L}_{un}$ is constructed as:
\begin{equation}
\mathcal{L}_{un} = \mathcal{L}_{un_1} + \mathcal{L}_{un_2}.
\end{equation}
The $\mathcal{L}_{un_1}$ (Equation~\ref{un1}) optimizes the error and uncertainty jointly, while the $\mathcal{L}_{un_2}$ (Equation~\ref{un2}) matches the distribution of uncertainty with the error.

\subsubsection{Adversarial Loss}
% \noindent \textbf{Adversarial Loss.}
Many researchers use GAN~\cite{mao2017least} to make synthesized talking face videos look more realistic~\cite{prajwal2020lip,zhong2023identity,wang2023seeing,zhang2023dinet}. Given its impressive performance, we also incorporate an adversarial loss:
\begin{equation}
\mathcal{L}_{ad} = \mathcal{L}_{G} + \mathcal{L}_{D},
\end{equation}
where $G$ and $D$ denote the generator and discriminator respectively, $\mathcal{L}_{G}$ and $\mathcal{L}_{D}$ can be represented as follows:
\begin{equation}
\begin{aligned}
&\mathcal{L}_G=\mathbb{E}(D(I_{gen})-1)^2,\\
&\mathcal{L}_D=\frac12(\mathbb{E}(D(I_{gt})-1)^2+\mathbb{E}(D(I_{gen}))^2),
\end{aligned}
\end{equation}
where $I_{gt}$ represents the ground truth image and $I_{gen}$ denotes the generated image. 

\subsubsection{Perception Loss}
% \noindent \textbf{Perception Loss.}
We minimize the perception loss between the generated image and the ground truth image,
\begin{equation}
\mathcal{L}_{pe}=\sum_i\left\|\phi_i({I_{gt}})-\phi_i(I_{gen})\right\|_1,
\end{equation}
where $\phi_i$ is the activation output of the $i$-th layer in the VGG-19~\cite{simonyan2014very} network.

\subsubsection{Lip-Sync Loss}
% \noindent \textbf{Lip-Sync Loss.}
We additionally train a syncnet~\cite{chung2017out} that comprises a visual encoder and an audio encoder. It determines the temporal alignment of visual and audio clips using contrastive loss.
\begin{equation}
\mathcal{L}_{sync}=\mathbb{E}(\mathrm{syncnet}(A,I_{gen})-1)^2,
\end{equation}
where $A$ denotes the audio clip.

The overall loss is the sum of the four losses:
\begin{equation}
\mathcal{L}= \lambda_1\mathcal{L}_{un} + \lambda_2\mathcal{L}_{ad} + \lambda_3 \mathcal{L}_{pe} + \lambda_4 \mathcal{L}_{sync}.
\end{equation}
In our experiments, we set $\lambda_1=1.0$, $\lambda_2=1.0$, $\lambda_3=10.0$, and $\lambda_4=0.1$.

\section{Experiments}

\subsubsection{Datasets}
% \noindent \textbf{Datasets.}
\label{datasets}
Two commonly used audio-visual talking face datasets, HDTF~\cite{zhang2021flow} and MEAD~\cite{wang2020mead}, are utilized in our experiments.

The HDTF dataset~\cite{zhang2021flow} contains approximately 400 in-the-wild videos that are collected with a resolution of either 720P or 1080P. For our experiments, we randomly select 10 of these videos for testing.

The MEAD dataset~\cite{wang2020mead} includes approximately 40 hours of emotional in-the-lab videos recorded in 1080P resolution. For our experiment, we are not specifically concern with emotional information, so we have chosen a total of 1,920 videos that depict neutral emotions and a frontal view. Additionally, we randomly select 50 videos from 5 different individuals for testing.

\subsubsection{Comparison Baselines}
% \noindent \textbf{Comparison Baselines.}
\label{cb}
We compare our method against various state-of-the-art talking face video generation methods, such as Wav2Lip~\cite{prajwal2020lip}, VideoReTalking~\cite{cheng2022videoretalking}, EAMM~\cite{ji2022eamm}, DINet~\cite{zhang2023dinet}, and TalkLip~\cite{zhang2023dinet}.

\subsubsection{Implementation Details}
% \noindent \textbf{Implementation Details.}
\label{id}
We implement and evaluate our JULNet using PyTorch on four RTX 3090 GPUs. During data processing, all videos are resampled at 25 fps. We crop the face region based on the 68 facial landmarks from OpenFace~\cite{baltruvsaitis2016openface} and resize all faces to a resolution of 416$\times$320. The mouth region covers a resolution of 256$\times$256 within the resized facial image. Since the HDTF dataset and MEAD dataset have limited subjects, we utilize a pre-trained DeepSpeech model~\cite{hannun2014deep} to extract audio features and improve generalization. The audio feature is then aligned with the video, also at 25 fps. For optimization, we employ the Adam optimizer~\cite{kingma2014adam} with a learning rate set at 1e-4.

\subsection{Quantitative Evaluation}

\subsubsection{Metrics}
% \noindent \textbf{Metrics.}
We utilize Peak Singnal-to-Noise Ration (PSNR) to assess the overall image quality and Learned Perceptual Image Patch Similarity (LPIPS)~\cite{zhang2018unreasonable} to evaluate the details.
Additionally, we employ \text{Fréchet} Inception Distance (FID)~\cite{heusel2017gans} to gauge image quality at the feature level. 
For evaluating lip synchronization, we introduce Lip Sync Error Distance (LSE-D) and Lip Sync Error Confidence (LSE-C), consistent with Wav2Lip~\cite{prajwal2020lip}, to assess the synchronization between lip movements and audio.

\subsubsection{Evaluation Results}
% \noindent \textbf{Evaluation Results.}
The quantitative evaluation results of the talking head video generation are illustrated in Table~\ref{tab1} and Table~\ref{tab2}. It is evident that our method outperforms other methods in terms of image quality across all aspects. This clearly demonstrates the positive impact of joint uncertainty learning. However, when it comes to the evaluation of audio-visual synchronization, our method falls short compared to Wav2Lip~\cite{prajwal2020lip}. One possible explanation is that the syncnet in Wav2Lip is trained on the LRS2 dataset, whereas our model is trained from scratch.

\begin{table}[t]
\resizebox{1\linewidth}{!}{
        \setlength{\tabcolsep}{1.0mm}
        \centering
        \begin{tabular}{lccccc}
        \toprule
        Methods & PSNR $\uparrow$ & LPIPS $\downarrow$ & FID $\downarrow$ & LSE-D $\downarrow$ & LSE-C $\uparrow$  \\
        Ground Truth  & N/A                        & 0              & 0            & 6.398             & 9.033                  \\ \midrule
        Wav2Lip       & 31.836  & 0.0391     &9.137       & \textbf{6.792}      & \textbf{8.877}   \\
        EAMM      & 29.371  & 0.0719     & 20.559           & 8.350      & 6.809   \\
        VideoReTalking     & 31.905  & \underline{0.0384}            & \underline{8.437}      & 7.109      & \underline{8.293}   \\
        DINet      & 31.824          & 0.0403         & 12.854    & 7.429      & 7.814             \\
        TalkLip        & \underline{32.003}   & 0.0393          & 9.974          & 8.836      & 6.533         \\
        \midrule
        
        JULNet (Ours) & \textbf{32.266}          & \textbf{0.0365} & \textbf{7.825}  & \underline{7.024}& {8.208}        \\ \bottomrule 
        \end{tabular}
    }
    \caption{Quantitative results of the talking head video generation on the HDTF dataset. The boldface indicates the best performance and the underline represents the second-best performance. 
    }
    \setlength{\abovecaptionskip}{0.25cm}
    \label{tab1}
    \vspace{-0.cm}
\end{table}

\begin{table}[t]
\resizebox{1\linewidth}{!}{
        \setlength{\tabcolsep}{1.0mm}
        \centering
        \begin{tabular}{lccccc}
        \toprule
        Methods & PSNR $\uparrow$ & LPIPS $\downarrow$ & FID $\downarrow$ & LSE-D $\downarrow$ & LSE-C $\uparrow$  \\
        Ground Truth  & N/A                        & 0              & 0            & 7.185             & 8.025                  \\ \midrule
        Wav2Lip      & 33.067  & 0.0242     &24.925      & \textbf{7.655}      & \underline{7.902}   \\
        EAMM       & 29.413  & 0.0883    & 50.301          & 8.416     & 6.351   \\
        VideoReTalking       & \underline{33.260}  & 0.0234            & \underline{23.350}      & 8.027      & \textbf{7.951}   \\
        DINet       & 33.026         & 0.0253         & 25.087    & 8.027      & 6.975             \\
        TalkLip       & 33.275   & \underline{0.0224}          & 30.071         &7.873      & 7.309        \\
        \midrule
        
        JULNet (Ours) & \textbf{33.817}          & \textbf{0.0210} & \textbf{19.525}  & \underline{7.801}& {7.147}        \\ \bottomrule 
        \end{tabular}
    }
    \setlength{\abovecaptionskip}{0.25cm}
    \caption{Quantitative results of the talking head video generation on the MEAD dataset. The boldface indicates the best performance and the underline represents the second-best performance. 
    }
    \label{tab2}
    \vspace{-0.cm}
\end{table}

\begin{figure*}
  \centering
  \includegraphics[width=0.9\linewidth]{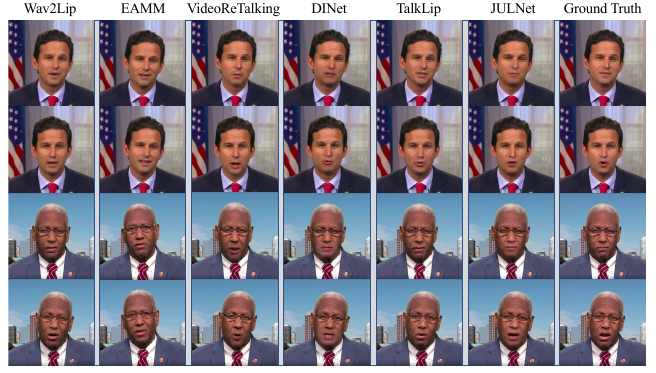}
  \caption{Qualitative comparison of talking face video generation by different methods on the HDTF dataset. JULNet has the best visual effect on lip movements and facial details. Please zoom in for better visualization.
  }
  \label{figure4}
% \vspace{-0.1cm}
\end{figure*}

% \begin{figure}
%   \centering
%   \includegraphics[width=\linewidth]{AnonymousSubmission/LaTeX/figure6.pdf}
%   \caption{User study results on three aspects, i.e. Image Quality, Video Realness, and Audio-Visual Synchronization.
%   }
%   \label{figure6}
% \vspace{-0.1cm}
% \end{figure}

\begin{figure*}
  \centering
  \includegraphics[width=0.9\linewidth]{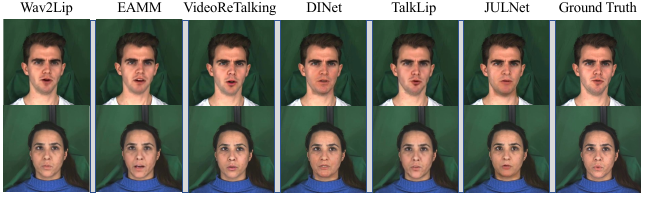}
  \caption{Qualitative comparison of talking face video generation by different methods on the MEAD dataset. JULNet has the best visual effect on lip movements and facial details. Please zoom in for better visualization.
  }
  \label{figure5}
% \vspace{-0.1cm}
\end{figure*}

\begin{figure}
  \centering
  \includegraphics[width=0.9\linewidth]{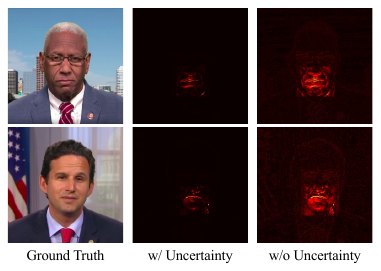}
  \caption{Comparison of the uncertainty maps generated by JULNet, with and without the inclusion of uncertainty loss. 
  }
  \label{figure7}
% \vspace{-0.1cm}
\end{figure}

\subsection{Qualitative Evaluation}
% \noindent \textbf{Evaluation Results.}
\subsubsection{Evaluation Results}
To evaluate image quality and lip synchronization more intuitively, we present a comparison of our method with others in Figure~\ref{figure4} and Figure~\ref{figure5}. We showcase key frames from a clip. The results demonstrate that our JULNet captures finer details and achieves the highest accuracy in lip synchronization. Its outputs notably align with the ground truth image. On the other hand, EAMM~\cite{ji2022eamm} and VideoReTalking~\cite{cheng2022videoretalking} tend to produce face images with moderate differences compared to the ground truth images. For instance, the eyes and lips are noticeably different from those in the ground truth images. Additionally, the results of Wav2Lip~\cite{wang2023seeing} and TalkLip~\cite{prajwal2020lip} appear slightly blurry.

\subsubsection{User Study}
\begin{figure*}
  \centering
  \includegraphics[width=0.9\linewidth]{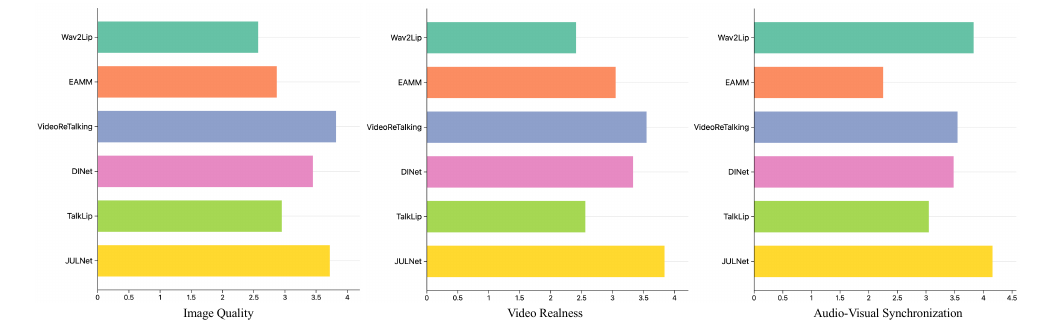}
  \caption{User study results on three aspects, i.e. Image Quality, Video Realness, and Audio-Visual Synchronization.
  }
  \label{figure6}
% \vspace{-0.1cm}
\end{figure*}

To conduct a more comprehensive evaluation of JULNet, we implement a user study questionnaire. We select 10 video clips that are generated during the quantitative evaluation and invite 16 participants to participate in the survey. The Mean Opinion Scores (MOS) rating protocol is utilized for this evaluation. Participants are required to rate the generated videos based on three aspects: (1) Image Quality, (2) Video Realness, and (3) Audio-Visual Synchronization. The average scores for each criterion are presented in Figure~\ref{figure6}. 
It is worth noting that although VideoReTalking~\cite{cheng2022videoretalking} scores about the same as our JULNet in terms of image quality, the images generated by VideoReTalking are more different from the ground truth images in terms of details, e.g., artifacts in the eyes and reddening of the lips.
The user study indicates that our JULNet is capable of producing videos of exceptional visual quality that achieve a high level of realism as perceived by humans.

\begin{table}[t]
    \resizebox{\linewidth}{!}{%
    \setlength{\tabcolsep}{1.0mm}
    \begin{tabular}{l|ccc|ccccc}
    \toprule
    \# & $\mathcal{L}_{pe}$ & $\mathcal{L}_{sync}$ & $\mathcal{L}_{un}$  
    &PSNR$\uparrow$ & LPIPS$\downarrow$ & FID$\downarrow$  &LSE-D$\downarrow$  &LSE-C$\uparrow$\\
    \midrule
    
    1  &-  &- &- 
     & 24.565 & 0.1229 & 17.081 & 9.396 & 6.020\\ 
    
    2  &\checkmark &- &- &31.939
    &0.0376 & 10.849 &8.982 &6.489\\ 

    3  &\checkmark &\checkmark &- &32.016
    &0.0380 & 10.254 &7.102 &8.071\\ 

    4  &\checkmark &\checkmark &\checkmark 
    &\textbf{32.266} & \textbf{0.0365} &\textbf{7.825} &\textbf{7.024} &\textbf{8.208}\\

    \bottomrule
    \end{tabular}
    }
    \caption{Ablation studies on loss functions. Best performance is highlighted in bold.}
    \label{tab3}
\end{table}

\begin{table}[t]
\centering
\resizebox{1\linewidth}{!}{
    \setlength{\tabcolsep}{2.5mm}
    \begin{tabular}{lccccc}
    \toprule
    Settings  & PSNR$\uparrow$ & LPIPS$\downarrow$  & FID$\downarrow$  & LSE-D$\downarrow$ & LSE-C$\uparrow$\\ \midrule
    $B=9$  & 32.124  & 0.0382  & 7.916  & 7.338 & 8.025   \\
    $B=10$  & 32.237 & 0.0370   & 8.014  & 7.309 & 8.058    \\
    $B=11$ &{32.266} & \textbf{0.0365} &\textbf{7.825} &\textbf{7.024} &\textbf{8.208}    \\
    $B=12$ & \textbf{32.290} & 0.0373   & 7.855  & 7.096 & 8.195    \\
    $B=13$ & 32.281 & 0.0373   & 7.914  & 7.127 & 8.129    \\
    \bottomrule
    w/o $\mathcal{L}_{un_1}$  & 32.015  & 0.0381  & 8.416  & 7.786 & 8.055   \\
    w/o $\mathcal{L}_{un_2}$  & 32.214 & 0.0369   & 8.002  & 7.379 & 8.017    \\
    JULNet  &\textbf{32.266} & \textbf{0.0365} &\textbf{7.825} &\textbf{7.024} &\textbf{8.208}    \\
    \bottomrule
    \end{tabular}
}
\caption{Ablation Study on uncertainty details. Best performance is highlighted in bold.}
\setlength{\abovecaptionskip}{0cm}
\label{tab4}
% \vspace{-0.5cm}
\end{table}

\begin{table}[t]
\resizebox{1\linewidth}{!}{
        \setlength{\tabcolsep}{0.5mm}
        \centering
        \begin{tabular}{lccccc}
        \toprule
        Methods & PSNR $\uparrow$ & LPIPS $\downarrow$ & FID $\downarrow$ & LSE-D $\downarrow$ & LSE-C $\uparrow$  \\
        Ground Truth  & N/A                        & 0              & 0            & 6.398             & 9.033                  \\ \midrule
        Wav2Lip       & 31.836  & 0.0391     &9.137       & {6.792}      & {8.877}   \\
        Wav2Lip + Uncertainty & \textbf{32.108}          & \textbf{0.0370} & \textbf{6.996}  & \textbf{6.680}& \textbf{8.893}        \\ \bottomrule 
        \end{tabular}
    }
    \caption{Quantitative results of the generalization experiment on the HDTF dataset. 
    }
    \label{sup_tab1}
    \vspace{-0.cm}
\end{table}

\subsection{Ablation Study}
We conduct an ablation study on the HDTF dataset to assess the contributions of various components in our JULNet.

\subsubsection{Loss Functions}
% \noindent \textbf{Loss Functions.}
We conduct ablation experiments on various loss functions, and the results are illustrated in Table~\ref{tab3}. The findings indicate that each loss function plays a critical role. Specifically, when comparing row 1 and row 2, the exclusion of perception loss results in a significant degradation of the face's realism and naturalness. It also becomes challenging to maintain the details and textures, leading to somewhat blurry generated images. Furthermore, by comparing row 2 and row 3, the inclusion of lip-sync loss ensures that the generated face video is synchronized with the input audio signal, resulting in substantial improvements in lip-sync metrics. Lastly, multi-task learning often leads to enhanced performance across all tasks. Therefore, by optimizing uncertainty, we can further improve image quality and audio-lip synchronization when comparing row 3 and row 4. 
Figure~\ref{figure7} provides a comparison of the uncertainty maps generated by JULNet, with and without the inclusion of uncertainty loss. This comparison clearly demonstrates that the use of uncertainty loss can greatly enhance the optimization of uncertainty, which is known to be positively correlated with the error.

\subsubsection{Uncertainty Details}
% \noindent \textbf{Uncertainty Details.}
In the previous part, the positive significance of uncertainty has been demonstrated. 
To further explore the details of uncertainty, additional ablation experiments are conducted. The results are presented in Table~\ref{tab4}.
To begin, we evaluate the number of bins ($B$) for the histogram. These bins are utilized to fit the error distribution and uncertainty distribution. We can obtain the best results when $B=11$, thus we set this setting for all experiments conducted.
Additionally, we perform ablation experiments on both parts of the uncertainty loss (Equation~\ref{un1} and Equation~\ref{un2}).
The results indicate that both components have a positive effect on the overall optimization process.
In this case, $\mathcal{L}_{un_1}$ is used to jointly optimize the error and uncertainty, while $\mathcal{L}_{un_2}$ is used to ensure that the distributions of the two can match, thus reflecting the error through uncertainty. 

\subsubsection{Generalization}
% The uncertainty strategy proposed in our paper can be used as a plug and play module for other methods. We conduct additional comparative experiments on Wav2Lip~\cite{prajwal2020lip}, and the experimental results are shown in Table~\ref{sup_tab1}, which proves the effectiveness of this strategy.
The uncertainty strategy proposed in our paper serves as a versatile solution that can be seamlessly integrated as a plug-and-play module into other existing methods to enhance their performance. To validate the practical effectiveness and generalization capability of this strategy, we conducted a series of comparative experiments on the classic lip synthesis model Wav2Lip~\cite{prajwal2020lip}, and the experimental results presented in Table~\ref{sup_tab1} clearly demonstrate that this strategy can significantly improve model performance, thoroughly confirming its effectiveness and practical value in real-world applications.

\section{Discussion}
\subsection{Ethical Consideration}
Our JULNet can produce highly realistic videos of talking faces. This capability has numerous advantages, but it also raises concerns about potential misuse. As part of our responsibility, we are committed to actively combating any malicious use of our technology. To fulfill this commitment, we will provide our generated videos to the Deepfake detection community. We firmly believe that the responsible application of our method will greatly contribute to the progress of digital human technology.

\subsection{Limitation and Future Work}
Our novel method successfully generates high-resolution talking face videos, setting a new standard in the field. However, despite the impressive results, our method still faces several challenges and limitations that need to be acknowledged and addressed in future work.

One significant limitation stems from our reliance on deforming reference images to inpaint the mouth region in the source face. While this technique enables realistic mouth movements, it struggles to handle complex, dynamic scenes with changeable lighting conditions, moving backgrounds, swaying earrings, flowing hair, or camera movements. In such cases, our method may produce visible artifacts, particularly when the mouth region overlaps with the background.

Moreover, our training data, sourced from the HDTF and MEAD datasets, predominantly consists of frontal-view videos. Consequently, our model's performance is constrained to a limited range of head poses, hindering its applicability to more diverse and unconstrained scenarios. Expanding the training data to include a wider variety of head orientations and camera angles can help mitigate this limitation and enhance the model's robustness.

While our method represents a significant step forward in high-resolution face visual dubbing, addressing these limitations and challenges will be crucial for future research. By tackling issues related to dynamic scenes, expanding the range of supported head poses, and enhancing the robustness to reference image selection, we can further push the baseline of realistic and convincing talking face.

\section{Conclusion}
\label{conclusion}
In this paper, we present JULNet, a novel taking face video generation method that incorporates a representation of uncertainty, which is intrinsically linked to visual error. Our method involves the design and implementation of a dedicated uncertainty module that predicts both the error map and the uncertainty map independently. This allows for a more comprehensive understanding of the model's confidence in its predictions and enables targeted improvements in areas with higher uncertainty.
% In this paper, we introduce JULNet, a new method that integrates a representation of uncertainty, which is directly tied to visual error. Our method involves designing an uncertainty module that predicts the error map and uncertainty map separately. 
Additionally, in order to align the uncertainty distribution with the error distribution, we employ a KL divergence term and utilize a histogram technique to approximate the distributions. Through the joint optimization of error and uncertainty, our JULNet outperforms previous methods in terms of generating high-fidelity talking face videos with audio-lip synchronization.

\bibliographystyle{ACM-Reference-Format}

% \appendix









\end{document}